\documentclass[conference]{IEEEtran}

\usepackage{blindtext}
\usepackage{eso-pic}
\IEEEoverridecommandlockouts
\usepackage{cite}
\usepackage{url}
\usepackage{booktabs}
\usepackage{pifont} 
\newcommand{\checkmark}{\ding{51}}
\usepackage{listings}
\usepackage{multirow}
\usepackage{amsmath,amssymb,amsfonts}
\usepackage{algorithmic}
\usepackage{graphicx}
\usepackage{textcomp}
\usepackage{xcolor}
\def\BibTeX{{\rm B\kern-.05em{\sc i\kern-.025em b}\kern-.08em
    T\kern-.1667em\lower.7ex\hbox{E}\kern-.125emX}}
    
\usepackage{eso-pic}

\lstdefinestyle{roadrlstyle}{
  basicstyle=\ttfamily\footnotesize,
  breaklines=true,
  breakatwhitespace=true,
  columns=flexible,
  keepspaces=true,
  frame=single,
  framerule=0.4pt,
  rulecolor=\color{black},
  numbers=none,
  showstringspaces=false,
  xleftmargin=0pt,
  xrightmargin=0pt,
  framexleftmargin=2pt,
  framexrightmargin=2pt,
  aboveskip=2pt,
  belowskip=2pt,
  tabsize=2
}

\begin{document}
\title{\vspace*{1cm} RoAd-RL: A Unified Library and Benchmark for Robust Adversarial Reinforcement Learning\\
% {\footnotesize \textsuperscript{*}Note: Sub-titles are not captured in Xplore and
% should not be used}
\thanks{This work was supported by Hightech-Agenda Bayern}
}

\author{\IEEEauthorblockN{1\textsuperscript{st} Adithya Mohan}
\IEEEauthorblockA{\textit{AImotion Bavaria} \\
\textit{Technische Hochschule Ingolstadt}\\
Ingolstadt, Germany \\
0009-0004-3572-9982}
\and
\IEEEauthorblockN{2\textsuperscript{nd} Daniel Kriegl}
\IEEEauthorblockA{\textit{AImotion Bavaria} \\
\textit{Technische Hochschule Ingolstadt}\\
Ingolstadt, Germany \\
0009-0001-1200-0011}
\and
\IEEEauthorblockN{3\textsuperscript{rd} Torsten Sch\"{o}n}
\IEEEauthorblockA{\textit{AImotion Bavaria} \\
\textit{Technische Hochschule Ingolstadt}\\
Ingolstadt, Germany \\
0000-0001-5763-3392}
}

\maketitle

\begin{abstract}
Deep Reinforcement Learning (DRL) has achieved significant success in robotics and autonomous systems, yet remains vulnerable to adversarial perturbations that can severely degrade performance. Research in adversarial reinforcement learning is often limited by fragmented implementations, inconsistent evaluation protocols, and poor reproducibility. To address these challenges, we present \textbf{RoAd-RL}, an open-source benchmarking framework that provides unified abstractions for policies, attacks, defenses, and robustness metrics, together with reproducible evaluation pipelines and seamless integration with Stable-Baselines3 and Gymnasium.

We evaluate DQN, PPO, and SAC agents in LunarLander and Highway-v0 under 192 attack-defense configurations. Results reveal substantial variations in robustness across environments and show that some commonly used defenses can be more detrimental than the attacks they aim to mitigate, while temporal smoothing consistently achieves strong performance. RoAd-RL establishes a standardized benchmark for adversarial reinforcement learning research and is publicly available at \url{https://pypi.org/project/road-rl}.
\end{abstract}


\begin{figure*}[t]
    \centering
    \includegraphics[width=\textwidth]{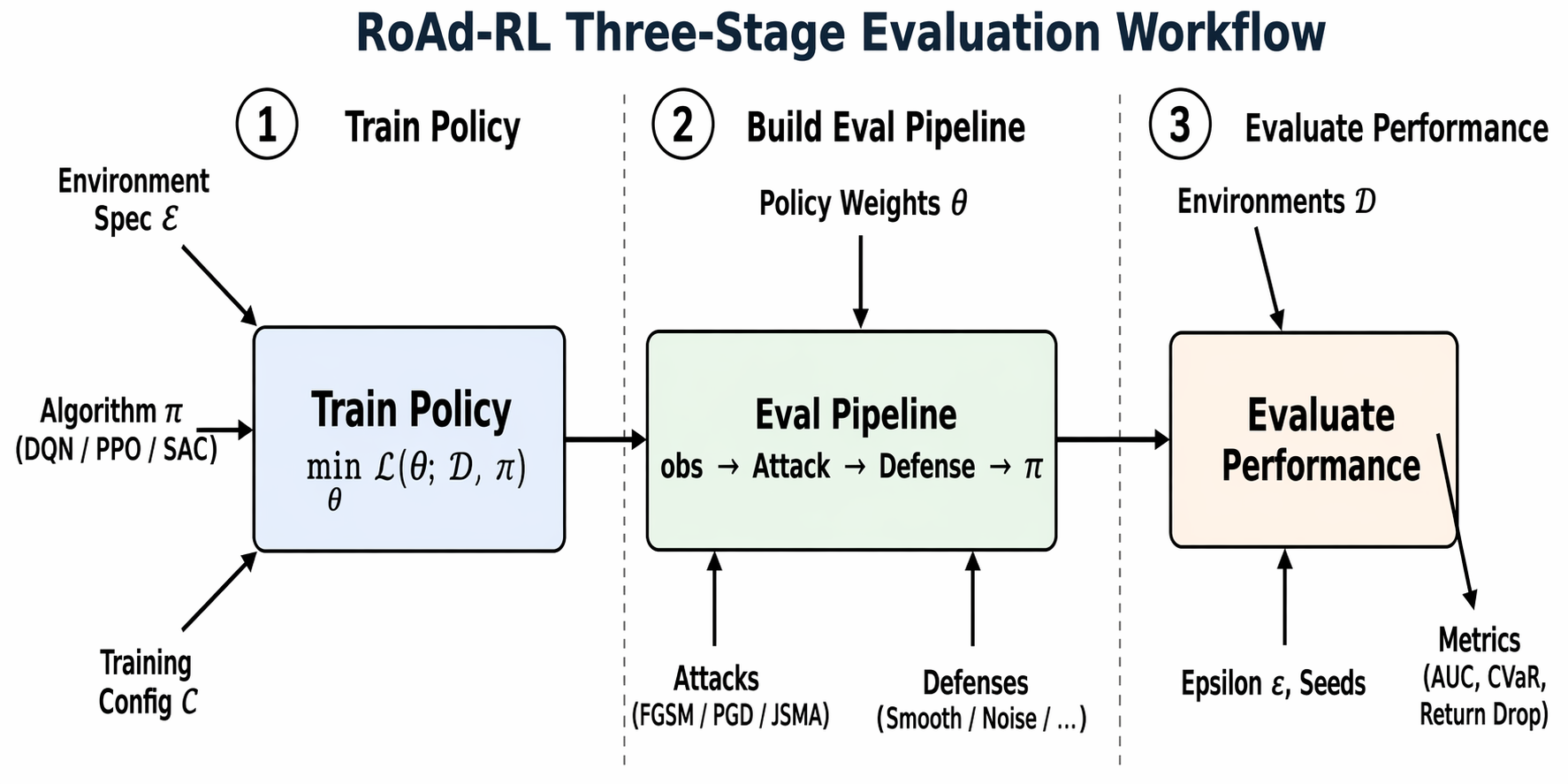}
    \caption{Overview of the proposed RoAd-RL framework. The workflow consists of three stages: policy training, attack-defense composition, and robustness evaluation. The modular architecture enables reproducible benchmarking across reinforcement learning algorithms, adversarial attacks, defense mechanisms, and evaluation metrics.}
    \label{fig:architecture}
\end{figure*}

%%%%%%%%%%%%%%%%%%%%%%%%%%%%%%%%%%%%%%%%%%%%%%%%%%%%%%%%%%
\section{Introduction}

Deep Reinforcement Learning (DRL) has achieved remarkable success in sequential decision-making tasks, including robotics, autonomous systems, intelligent transportation, and industrial control \cite{mohan2026toward}. By combining reinforcement learning with deep neural networks, DRL agents can learn complex behaviors directly from interactions with their environment \cite{karpenahalli2025evolution}. However, similar to supervised learning models, DRL policies remain vulnerable to adversarial perturbations that can significantly degrade performance and induce unsafe behavior \cite{huang2017adversarial,zhang2020robust}. These vulnerabilities pose serious challenges for deploying DRL in safety-critical applications \cite{mohan2025advancing, mohan2026real}.

To address these weaknesses, numerous adversarial attacks and defense mechanisms have been proposed. Existing attacks include gradient-based approaches such as FGSM, PGD, and JSMA, as well as more advanced optimization-based strategies \cite{goodfellow2014explaining,madry2017towards,papernot2016crafting,sun2020stealthy}. Likewise, defenses such as adversarial training, feature squeezing, temporal smoothing, randomized smoothing, and anomaly detection have been extensively studied \cite{zhang2020robust,xu2017feature,cohen2019certified}. Despite this growing body of work, robustness evaluation in reinforcement learning remains fragmented due to custom implementations, inconsistent evaluation protocols, and limited reproducibility.

In supervised learning, frameworks such as RobustBench \cite{croce2020robustbench}, Foolbox \cite{rauber2017foolbox}, and CleverHans \cite{papernot2018cleverhans} have greatly improved reproducibility by providing standardized attacks, defenses, and evaluation procedures. In contrast, adversarial reinforcement learning lacks a widely adopted benchmarking framework that offers comparable extensibility and reproducibility.

To bridge this gap, we present \textbf{RoAd-RL}, a unified benchmarking and evaluation framework for Robust Adversarial Reinforcement Learning. RoAd-RL introduces a modular architecture based on four reusable abstractions: \texttt{Policy}, \texttt{Attack}, \texttt{Defense}, and \texttt{Metric}. Through integration with Stable-Baselines3 \cite{raffin2021stable} and Gymnasium \cite{towers2024gymnasium}, the framework enables reproducible robustness evaluation across multiple algorithms and environments.

Figure~\ref{fig:architecture} illustrates the overall RoAd-RL workflow. First, a reinforcement learning policy is trained within a selected environment. Second, adversarial attacks and defense mechanisms are composed into a unified evaluation pipeline. Finally, robustness metrics are computed across multiple perturbation budgets and random seeds to produce standardized benchmark results. This design allows new attacks, defenses, and policies to be evaluated without modifying the underlying infrastructure.

The main contributions of this work are:

\begin{itemize}
    \item We present \textbf{RoAd-RL}, a unified benchmarking and evaluation framework for adversarial reinforcement learning.
    
    \item We introduce a modular architecture based on \texttt{Policy}, \texttt{Attack}, \texttt{Defense}, and \texttt{Metric} abstractions for extensible and reproducible robustness evaluation.
    
    % \item We provide implementations of widely used attacks, defenses, adversarial training utilities, and standardized robustness metrics within a common framework.
    
    % \item We establish a benchmark comprising 192 attack-defense configurations across multiple reinforcement learning algorithms and environments.
\end{itemize}

The remainder of this paper is organized as follows. Section II reviews related work. Section III presents the RoAd-RL Framework Design. Section IV describes the benchmark methodology and experimental setup. Finally, Section V concludes the paper and outlines future directions.
%%%%%%%%%%%%%%%%%%%%%%%%%%%%%%%%%%%%%%%%%%%%%%%%%%%%%%%%%%

\section{Related Work}

Adversarial robustness in reinforcement learning has become an increasingly important research area due to the growing deployment of learning-based agents in safety-critical domains such as autonomous systems, robotics, intelligent transportation, and industrial automation. Existing research can generally be categorized into three major directions: adversarial attacks on reinforcement learning agents, defense mechanisms for improving robustness, and benchmarking frameworks for robustness evaluation.

\subsection{Adversarial Attacks on Reinforcement Learning}

The vulnerability of reinforcement learning policies to adversarial perturbations was first systematically demonstrated by Huang et al.~\cite{huang2017adversarial}. Their work showed that policies trained using Deep Q-Networks (DQN), Trust Region Policy Optimization (TRPO), and Asynchronous Advantage Actor-Critic (A3C) could be significantly degraded through carefully crafted adversarial observations generated using the Fast Gradient Sign Method (FGSM)~\cite{goodfellow2014explaining}. The study established that even small perturbations can substantially alter policy behavior and reduce cumulative rewards.

Subsequent work extended adversarial attacks to stronger optimization-based methods. Pattanaik et al.~\cite{pattanaik2018robust} adapted Projected Gradient Descent (PGD)~\cite{madry2017towards} to reinforcement learning and demonstrated that iterative perturbation strategies often outperform single-step attacks. Jacobian-based Saliency Map Attacks (JSMA)~\cite{papernot2016crafting} have also been applied to reinforcement learning settings, highlighting that sparse perturbations targeting only a subset of observation features can significantly affect policy decisions.

More recently, researchers have explored advanced threat models that exploit the sequential nature of reinforcement learning. Sun et al.~\cite{sun2020stealthy} proposed temporally consistent attacks that maintain perturbation coherence across multiple time steps, making them more difficult to detect while preserving attack effectiveness. Additional studies have investigated reward poisoning, environment manipulation, policy poisoning, and adversarial policy training, demonstrating that vulnerabilities can emerge throughout the reinforcement learning pipeline.

% Collectively, these studies demonstrate that adversarial vulnerabilities represent a fundamental challenge for reinforcement learning systems rather than a weakness of a specific algorithm or environment.

\subsection{Defense Mechanisms}

To mitigate adversarial vulnerabilities, a wide range of defense strategies has been proposed. Adversarial training remains one of the most widely adopted approaches and involves exposing the agent to adversarially perturbed observations during training \cite{madry2017towards,zhang2020robust}. Although effective in many settings, adversarial training typically incurs substantial computational overhead and may exhibit limited generalization to unseen attacks.

Inference-time defenses provide an alternative that does not require policy retraining. Observation preprocessing techniques such as normalization, clipping, and feature squeezing attempt to remove adversarial perturbations before policy inference \cite{xu2017feature}. Temporal smoothing approaches leverage the temporal structure of reinforcement learning by averaging observations over multiple time steps, thereby suppressing transient perturbations \cite{lin2017tactics}. Randomized smoothing techniques introduce controlled stochasticity and can provide probabilistic robustness guarantees under certain assumptions \cite{lecuyer2019certified,cohen2019certified}.

More sophisticated approaches rely on anomaly detection and adversarial detection mechanisms to identify malicious observations before they reach the policy network. While promising, existing defenses often exhibit strong environment dependency and can introduce significant performance degradation when deployed outside their intended settings. Consequently, no universally effective defense has yet emerged for adversarial reinforcement learning.

\subsection{Robustness Evaluation Frameworks}

The supervised learning community has benefited greatly from standardized robustness evaluation platforms such as RobustBench~\cite{croce2020robustbench}, Foolbox~\cite{rauber2017foolbox}, and CleverHans~\cite{papernot2018cleverhans}. These frameworks provide common implementations of attacks, defenses, datasets, and evaluation protocols, thereby enabling reproducible experimentation and fair comparison between competing methods.

In contrast, adversarial reinforcement learning research remains highly fragmented. Most studies rely on custom implementations and independently developed evaluation pipelines that differ substantially in attack configurations, perturbation budgets, evaluation metrics, and reporting methodologies. As a result, robustness claims are often difficult to reproduce and compare across studies.

% Several research repositories provide implementations of specific attacks or defenses. However, these repositories are typically designed for individual publications and rarely offer reusable benchmarking infrastructure capable of supporting standardized robustness evaluation across multiple algorithms and environments.

\subsection{Positioning of RoAd-RL}

The limitations of existing robustness evaluation practices motivate the development of RoAd-RL. Unlike prior attack-specific or defense-specific repositories, RoAd-RL provides a unified benchmarking and evaluation framework that standardizes the interaction between policies, attacks, defenses, and robustness metrics. The framework emphasizes reproducibility, extensibility, and interoperability while integrating seamlessly with widely adopted reinforcement learning ecosystems such as Stable-Baselines3~\cite{raffin2021stable} and Gymnasium~\cite{towers2024gymnasium}.

Table~\ref{tab:roadrl_comparison} compares RoAd-RL against existing adversarial robustness frameworks and reinforcement learning robustness repositories. Unlike prior solutions, RoAd-RL combines attack implementations, defense mechanisms, standardized metrics, reproducible evaluation protocols, command-line execution, Stable-Baselines3 integration, Gymnasium compatibility, and pip-installable deployment within a unified framework.

%%%%%%%%%%%%%%%%%%%%%%%%%%%%%%%%%%%%%%%%%%%%%%%%%%%%%%%%%%
\begin{table*}[t]
\centering
\caption{Comparison of RoAd-RL with existing adversarial robustness libraries and DRL robustness repositories. Unlike prior frameworks, RoAd-RL provides a unified DRL-focused benchmark pipeline with attacks, defenses, metrics, reproducibility support, Stable-Baselines3 integration, Gymnasium compatibility, and both CLI- and Python-based execution.}
\label{tab:roadrl_comparison}
\small
\begin{tabular}{lccccccccc}
\toprule
Framework & DRL & Attacks & Defenses & Metrics & CLI & Reproducible Sweeps & SB3 & Gymnasium & Pip Install \\
\midrule
RobustBench~\cite{croce2020robustbench} & $\times$ & \checkmark & \checkmark & \checkmark & $\times$ & $\times$ & $\times$ & $\times$ & \checkmark \\
Foolbox~\cite{rauber2017foolbox} & Limited & \checkmark & $\times$ & Limited & $\times$ & $\times$ & $\times$ & $\times$ & \checkmark \\
CleverHans~\cite{papernot2018cleverhans} & $\times$ & \checkmark & Limited & Limited & $\times$ & $\times$ & $\times$ & $\times$ & \checkmark \\
RL-Adv~\cite{zhang2020robust} & \checkmark & \checkmark & \checkmark & Limited & $\times$ & $\times$ & $\times$ & $\times$ & $\times$ \\
StateAdvDRL~\cite{pattanaik2018robust} & \checkmark & \checkmark & Limited & $\times$ & $\times$ & $\times$ & $\times$ & $\times$ & $\times$ \\
RoAd-RL (ours) & \checkmark & \checkmark & \checkmark & \checkmark & \checkmark & \checkmark & \checkmark & \checkmark & \checkmark \\
\bottomrule
\end{tabular}
\end{table*}

%%%%%%%%%%%%%%%%%%%%%%%%%%%%%%%%%%%%%%%%%%%%%%%%%%%%%%%%%%
\section{RoAd-RL Framework Design}
\label{sec:library}

Adversarial robustness research in reinforcement learning has grown rapidly over the past decade, resulting in a diverse collection of attack algorithms, defense mechanisms, evaluation metrics, and experimental protocols. However, most existing studies rely on custom implementations tightly coupled to specific environments, reinforcement learning algorithms, and threat models. As a result, reproducing published results and performing fair comparisons across different robustness methods remains challenging. RoAd-RL is designed to address this limitation by providing a unified framework that standardizes robustness evaluation while maintaining flexibility for future extensions.

The framework follows three fundamental design principles. First, \emph{composability} enables attacks, defenses, policies, and metrics to operate as independent components that can be freely combined within a common evaluation pipeline. Second, \emph{extensibility} ensures that all major components inherit from abstract base classes, allowing researchers to integrate new methods through subclassing without modifying framework internals. Third, \emph{reproducibility} is achieved through deterministic seed management, structured experiment logging, and standardized evaluation procedures, ensuring that robustness experiments can be reproduced consistently across different computing environments.

%%%%%%%%%%%%%%%%%%%%%%%%%%%%%%%%%%%%%%%%%%%%%%%%%%%%%%%%%%

\subsection{Architecture Overview}

RoAd-RL organizes adversarial robustness evaluation into three distinct stages, as illustrated in Fig.~\ref{fig:architecture}.

\textbf{Stage 1: Policy Training.}
A reinforcement learning policy $\pi_{\boldsymbol{\theta}}$ is trained using a selected learning algorithm and environment. RoAd-RL currently leverages Stable-Baselines3~\cite{raffin2021stable} as the primary backend and supports DQN, PPO, and SAC implementations through dedicated policy adapters.

\textbf{Stage 2: Attack--Defense Composition.}
The trained policy is combined with an adversarial attack and an optional defense mechanism to construct a robustness evaluation pipeline. During evaluation, observations generated by the environment are first perturbed by the attack module and subsequently processed by the defense before being forwarded to the policy network. This process can be expressed as

\[
\text{obs}
\xrightarrow{\varepsilon}
\texttt{Attack}
\rightarrow
\texttt{Defense}
\rightarrow
\pi
\rightarrow
\text{action}.
\]

This modular composition enables arbitrary combinations of attacks and defenses without requiring modifications to the policy implementation.

\textbf{Stage 3: Robustness Evaluation.}
The resulting pipeline is evaluated across multiple perturbation budgets, random seeds, and episodes. Performance statistics are aggregated into standardized robustness metrics that quantify the impact of adversarial perturbations and the effectiveness of defense mechanisms.

%%%%%%%%%%%%%%%%%%%%%%%%%%%%%%%%%%%%%%%%%%%%%%%%%%%%%%%%%%

\subsection{Core Framework Abstractions}

The framework is built around four core abstractions: \texttt{Policy}, \texttt{Attack}, \texttt{Defense}, and \texttt{Metric}. Together, these abstractions decouple learning algorithms from robustness evaluation and enable reusable benchmarking workflows.

\subsubsection{Policy}

The \texttt{Policy} abstraction provides a unified interface for reinforcement learning agents independent of the underlying learning algorithm. Every policy implements a single action-selection method,

\begin{lstlisting}[style=roadrlstyle]
act(obs) -> action
\end{lstlisting}

which allows attacks, defenses, and evaluation pipelines to interact with policies consistently across different algorithms.

For gradient-based attacks, RoAd-RL additionally provides a \texttt{DifferentiablePolicy} interface exposing

\begin{lstlisting}[style=roadrlstyle]
forward(obs)
loss(obs)
\end{lstlisting}

thereby enabling backpropagation through the policy network. The framework currently provides adapters for Stable-Baselines3 policies, including \texttt{SB3DQNPolicy}, \texttt{SB3PPOPolicy}, and \texttt{SB3SACPolicy}, as well as native PyTorch wrappers.

\subsubsection{Attack}

The \texttt{Attack} abstraction encapsulates adversarial perturbation generation and defines a common interface

\begin{lstlisting}[style=roadrlstyle]
apply(obs, policy, ctx) -> obs
\end{lstlisting}

where \texttt{ctx} is a \texttt{StepContext} object containing the perturbation budget, evaluation step, and random seed. This design separates perturbation generation from environment-specific logic and enables attacks to be reused across multiple reinforcement learning algorithms.

Gradient-based attacks inherit from \texttt{GradientAttack}, which requires access to a \texttt{DifferentiablePolicy}. Non-gradient methods such as random noise attacks and future black-box attacks can be integrated without modification to the remaining framework components.

\subsubsection{Defense}

The \texttt{Defense} abstraction represents inference-time robustness mechanisms that operate exclusively on observations prior to policy inference. Each defense implements

\begin{lstlisting}[style=roadrlstyle]
apply(obs, ctx) -> obs
\end{lstlisting}

and returns a transformed observation that is subsequently processed by the policy network. By preventing direct access to policy internals, the framework ensures that defenses remain independently testable and fully composable with arbitrary attacks.

\subsubsection{Metric}

The \texttt{Metric} abstraction standardizes robustness reporting through a common interface

\begin{lstlisting}[style=roadrlstyle]
compute(episodes) -> MetricResult
\end{lstlisting}

where \texttt{MetricResult} contains a metric name, scalar value, and auxiliary information. RoAd-RL currently supports four categories of evaluation metrics:

\begin{itemize}
    \item \textbf{Return Metrics}: Mean return, median return, and normalized performance degradation.
    \item \textbf{Risk Metrics}: Conditional Value at Risk (CVaR) and worst-percentile return.
    \item \textbf{Safety Metrics}: Episode termination rates and environment-specific safety violations.
    \item \textbf{Robustness Metrics}: Area Under the Curve (AUC--$\varepsilon$) and perturbation sensitivity measures.
\end{itemize}

%%%%%%%%%%%%%%%%%%%%%%%%%%%%%%%%%%%%%%%%%%%%%%%%%%%%%%%%%%

\subsection{Framework Usage}

One of the primary objectives of RoAd-RL is to reduce the implementation effort required to conduct adversarial robustness experiments. Listing~\ref{lst:usage} demonstrates a complete evaluation pipeline consisting of a trained DQN policy, a PGD attack, and a temporal smoothing defense.

\begin{figure}[t]
\centering
\begin{minipage}{\columnwidth}
\begin{lstlisting}[style=roadrlstyle,caption={Minimal Python usage example.},label={lst:usage}]
from road_rl.policies import SB3DQNPolicy
from road_rl.attacks import PGDAttack
from road_rl.defenses import MovingAverageSmoothingDefense
from road_rl.eval import Evaluator, EvalSpec
from stable_baselines3 import DQN

policy = SB3DQNPolicy(DQN.load("lunar_dqn.zip"))
attack = PGDAttack(steps=10, norm="linf")
defense = MovingAverageSmoothingDefense(window=3)

spec = EvalSpec(env_id="LunarLander-v2",
                algorithm="dqn",
                epsilons=[0.0, 0.01, 0.025, 0.05],
                seeds=[0, 1, 2],
                attack=attack,
                defense=defense)
results = Evaluator().run(policy, [spec])
\end{lstlisting}
\end{minipage}
\end{figure}

The modular architecture allows attacks and defenses to be exchanged without modifying the surrounding evaluation workflow, thereby facilitating systematic attack-defense benchmarking.

%%%%%%%%%%%%%%%%%%%%%%%%%%%%%%%%%%%%%%%%%%%%%%%%%%%%%%%%%%

\subsection{Implemented Adversarial Attacks}

RoAd-RL currently provides implementations of three widely adopted gradient-based adversarial attacks. All attacks operate in observation space and maximize policy loss with respect to the current observation. For discrete-action policies, perturbations are constrained using the $\ell_{\infty}$ norm, while continuous-action policies employ $\ell_2$ constraints.

\textbf{Fast Gradient Sign Method (FGSM)}~\cite{goodfellow2014explaining} generates a single-step perturbation according to

\[
\tilde{o}
=
o
+
\varepsilon
\cdot
\operatorname{sign}
\left(
\nabla_o \mathcal{L}
(\pi_\theta,o)
\right).
\]

\textbf{Projected Gradient Descent (PGD)}~\cite{madry2017towards} extends FGSM through iterative optimization. RoAd-RL uses a default configuration of $K=10$ optimization steps with step size $\alpha=\varepsilon/K$ and optional random initialization.

\textbf{Jacobian-based Saliency Map Attack (JSMA)}~\cite{papernot2016crafting} constructs sparse perturbations by modifying only the most influential observation dimensions according to a Jacobian-based saliency ranking. By default, the top-$k=2$ features are perturbed during each iteration.

%%%%%%%%%%%%%%%%%%%%%%%%%%%%%%%%%%%%%%%%%%%%%%%%%%%%%%%%%%

\subsection{Implemented Defense Mechanisms}

RoAd-RL currently provides seven inference-time defenses, all implementing the common \texttt{Defense.apply(obs, ctx)} interface.

\begin{itemize}
    \item \textbf{NormalizeClip}: Projects observations to the interval $[-1,1]$ using environment bounds.
    
    \item \textbf{Temporal Smoothing}: Applies a moving average over the most recent $w=3$ observations~\cite{lin2017tactics}.
    
    \item \textbf{Gaussian Noise}: Injects $\mathcal{N}(0,\sigma^2)$ noise with $\sigma=0.01$ to perform randomized smoothing~\cite{lecuyer2019certified}.
    
    \item \textbf{Feature Squeezing}: Quantizes observations to $b=4$ bits~\cite{xu2017feature}.
    
    \item \textbf{Median Smoothing}: Computes the per-feature median across a sliding window of $w=5$ observations.
    
    \item \textbf{Outlier Clip (MAD)}: Uses robust statistics to identify and clip anomalous observations beyond $k=4$ median absolute deviations.
    
    \item \textbf{Adversarial Detector}: Detects suspicious observations using rolling-buffer statistics and replaces flagged inputs with the most recent clean observation.
\end{itemize}

Together, these defenses span preprocessing-based, smoothing-based, statistical, and detection-based robustness strategies commonly investigated in adversarial machine learning.

%%%%%%%%%%%%%%%%%%%%%%%%%%%%%%%%%%%%%%%%%%%%%%%%%%%%%%%%%%

\subsection{Evaluation Pipeline and Command-Line Interface}

The \texttt{Evaluator} module orchestrates attack-defense sweeps across perturbation budgets and random seeds. At every interaction step, a \texttt{StepContext} object is passed to both attack and defense modules, ensuring deterministic execution and reproducible evaluation.

Experimental outcomes are aggregated into \texttt{ExperimentResult} objects and stored in structured CSV and JSON formats suitable for downstream analysis and visualization. Episode-level seeds are generated deterministically according to

\begin{equation}
s_\mathrm{ep}
=
(s \times 10007 + \ell)
\bmod 2^{31}
\end{equation}

where $s$ denotes the experiment seed and $\ell$ represents the episode index.

RoAd-RL additionally provides a command-line interface for executing robustness evaluations without requiring custom Python scripts.

\begin{figure}[t]
\centering
\begin{minipage}{\columnwidth}
\begin{lstlisting}[style=roadrlstyle,caption={CLI-based evaluation.},label={lst:cli}]
road-rl eval --env-id LunarLander-v2 \
  --algorithm dqn \
  --attack pgd \
  --defense smoothing \
  --eps 0.0 0.01 0.025 0.05 \
  --seeds 0 1 2 \
  --out results/
\end{lstlisting}
\end{minipage}
\end{figure}

%%%%%%%%%%%%%%%%%%%%%%%%%%%%%%%%%%%%%%%%%%%%%%%%%%%%%%%%%%

\subsection{Extensibility}

A central design objective of RoAd-RL is to support rapid experimentation with new robustness methods. New attacks, defenses, policies, and metrics can be integrated by subclassing the corresponding abstract base class and implementing the required interface.

\begin{figure}[t]
\centering
\begin{minipage}{\columnwidth}
\begin{lstlisting}[style=roadrlstyle,caption={Custom defense example.},label={lst:extend}]
from road_rl.defenses.base import Defense

class MyDefense(Defense):
    def apply(self, obs, ctx):
        return obs * 0.99
\end{lstlisting}
\end{minipage}
\end{figure}

Once implemented, the new component can immediately participate in any evaluation pipeline without modification to the framework core. This design enables straightforward extension of RoAd-RL and encourages community-driven development of robustness evaluation methodologies.

%%%%%%%%%%%%%%%%%%%%%%%%%%%%%%%%%%%%%%%%%%%%%%%%%%%%%%%%%%
\section{Benchmark Methodology and Experimental Evaluation}
\label{sec:benchmark}

To validate RoAd-RL and establish reproducible robustness baselines, we evaluate the complete attack--defense grid across two reinforcement learning environment families. The benchmark is designed not only to demonstrate the functionality of the framework, but also to analyze how adversarial attacks and inference-time defenses interact across different policy classes and environment dynamics.

%%%%%%%%%%%%%%%%%%%%%%%%%%%%%%%%%%%%%%%%%%%%%%%%%%%%%%%%%%

\subsection{Experimental Setup}

\textbf{Environments.}
We evaluate RoAd-RL on two environment families with distinct observation structures and control characteristics. The first is \texttt{LunarLander-v2}~\cite{towers2024gymnasium,sekaran2025urbaning}, which provides an 8-dimensional observation space and supports both discrete and continuous control variants. The second is \texttt{highway-v0}~\cite{leurent2018environment,rossle2026driving}, a structured driving environment with a 25-dimensional kinematic observation representation and discrete or continuous control settings. These environments were selected because they expose different robustness characteristics: LunarLander represents a compact control task with sensitive dynamics, while Highway-v0 represents a structured decision-making task with semantically meaningful kinematic features.

\textbf{Agents.}
We train three widely used reinforcement learning algorithms: Deep Q-Networks (DQN)~\cite{mnih2015human}, Proximal Policy Optimization (PPO)~\cite{schulman2017proximal}, and Soft Actor-Critic (SAC)~\cite{haarnoja2018soft}. These algorithms represent value-based, on-policy policy-gradient, and off-policy actor-critic learning paradigms, respectively. All agents are trained using Stable-Baselines3 with default hyperparameters. Table~\ref{tab:training} summarizes the best training performance obtained using a 200-episode moving window.

\begin{table}[t]
\caption{Trained agent performance measured using the best 200-episode reward window.}
\label{tab:training}
\centering
\footnotesize
\begin{tabular}{llll}
\toprule
\textbf{Environment} & \textbf{Algo} & \textbf{Episodes} & \textbf{Best Reward} \\
\midrule
LunarLander-v2           & DQN & 4{,}227 & 273.4 \\
LunarLander-v2           & PPO & 2{,}328 & 122.5 \\
LunarLanderContinuous-v2 & SAC & 3{,}089 & 286.0 \\
\midrule
Highway-v0               & DQN & 4{,}087 & 19.1 \\
Highway-v0               & PPO & 2{,}898 & 28.1 \\
Highway-v0               & SAC & 2{,}376 & 163.5 \\
\bottomrule
\end{tabular}
\end{table}

\begin{figure}[t]
  \centering
  \includegraphics[width=\columnwidth]{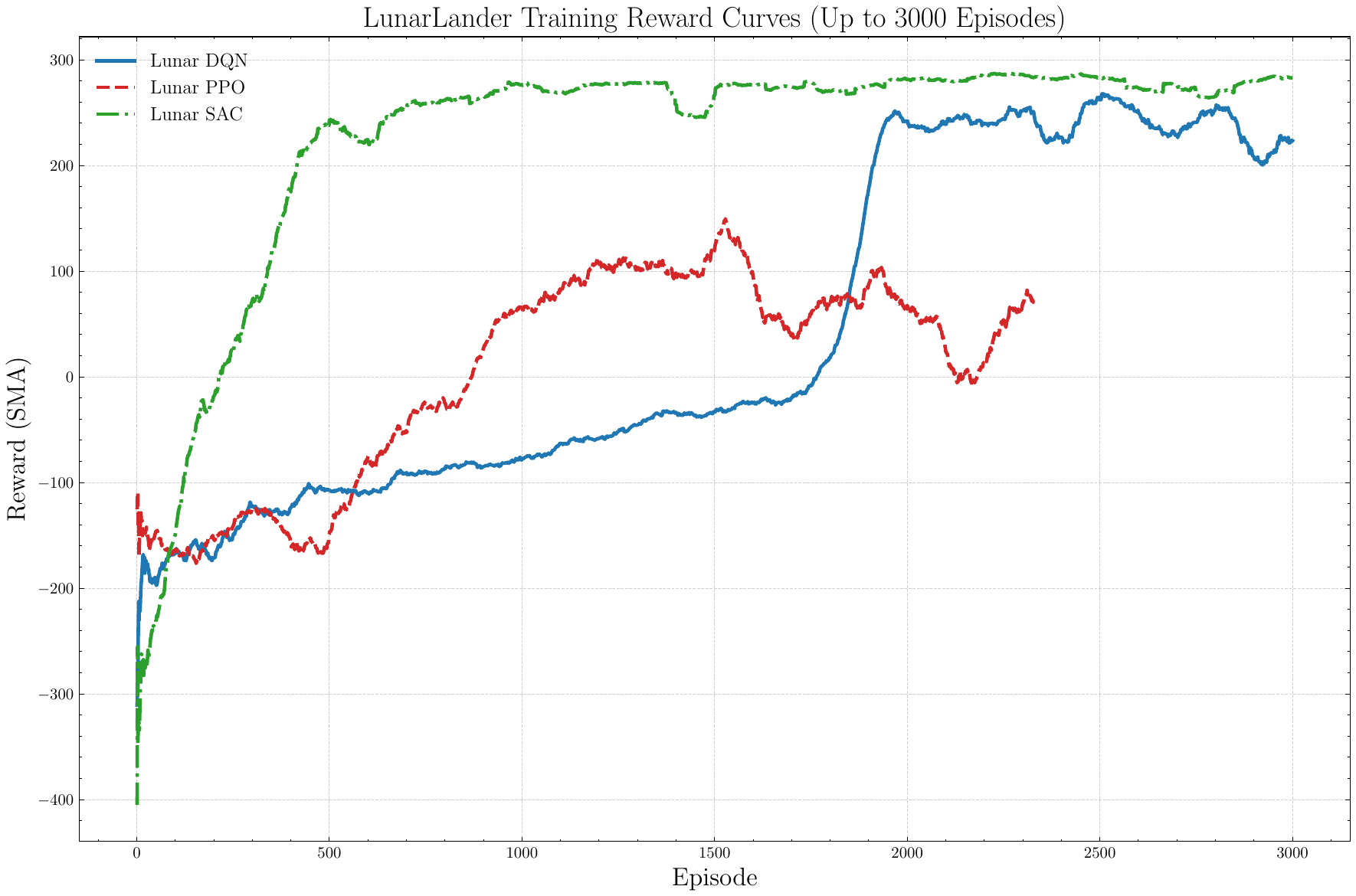}
  \caption{Smoothed training reward curves for DQN, PPO, and SAC on LunarLander-v2/Continuous-v2.}
  \label{fig:lunar_training}
\end{figure}

\begin{figure}[t]
  \centering
  \includegraphics[width=\columnwidth]{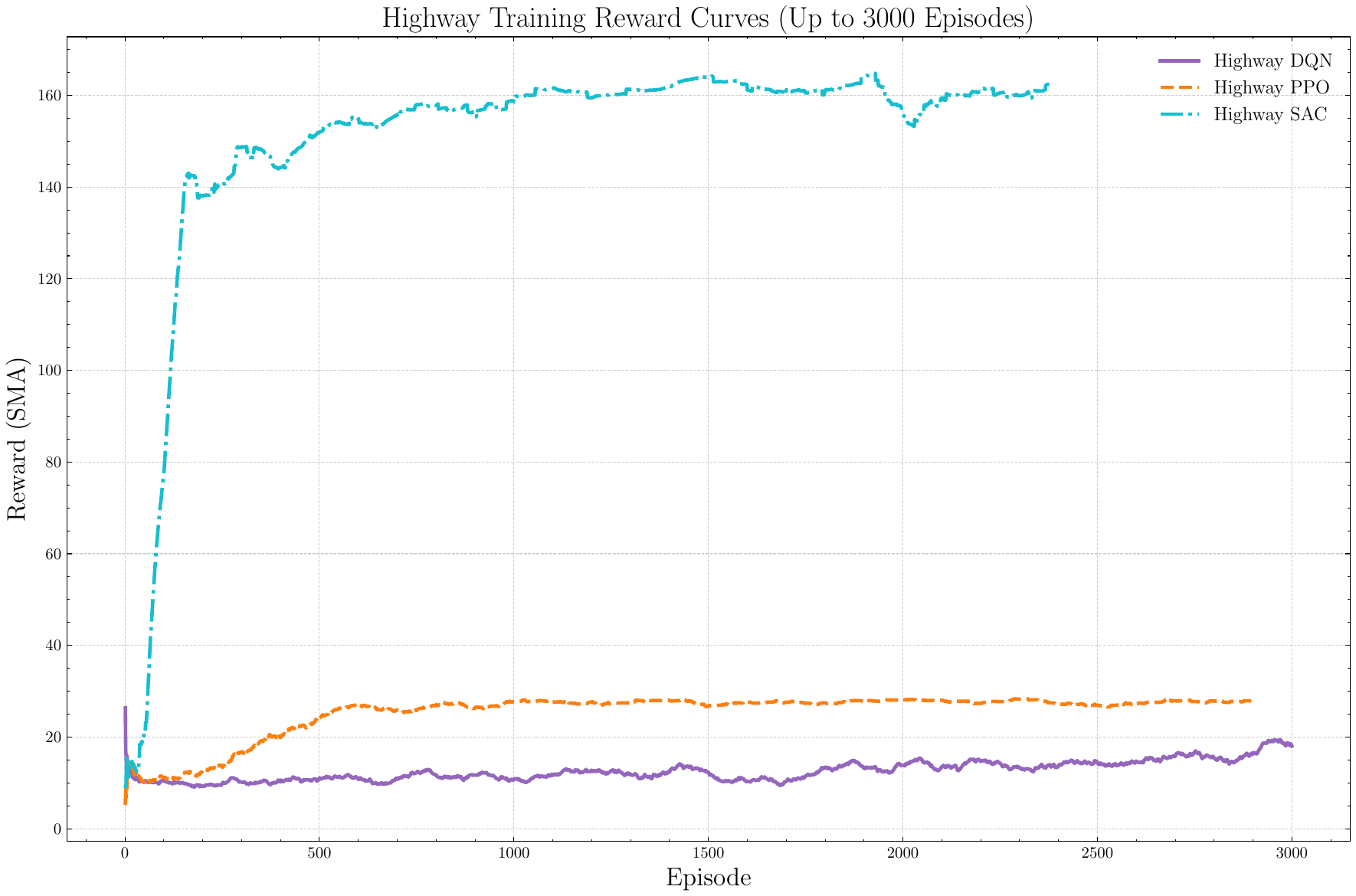}
  \caption{Smoothed training reward curves for DQN, PPO, and SAC on Highway-v0.}
  \label{fig:highway_training}
\end{figure}

Fig.~\ref{fig:lunar_training} and Fig.~\ref{fig:highway_training} show the smoothed training reward curves for the evaluated agents. These curves provide the clean-performance reference point used before adversarial evaluation. On LunarLander, SAC achieves the highest peak reward, while DQN reaches strong performance in the discrete-action setting. PPO plateaus at a lower reward level, indicating a weaker baseline policy. On Highway-v0, SAC obtains the highest overall reward, while DQN and PPO converge to lower but stable reward regimes.

%%%%%%%%%%%%%%%%%%%%%%%%%%%%%%%%%%%%%%%%%%%%%%%%%%%%%%%%%%

\subsection{Evaluation Protocol}

For each trained agent, we evaluate the interaction between adversarial attacks and inference-time defenses using a full factorial benchmark. The evaluated attacks are FGSM, PGD, and JSMA, together with the clean baseline. The evaluated defenses include no defense, temporal smoothing, median smoothing, Gaussian noise, feature squeezing, outlier clipping, and adversarial detection.

For each of the $6 \times 4 \times 8 = 192$ agent--attack--defense configurations, we sweep over perturbation budgets
\[
\varepsilon \in \{0.0, 0.005, 0.01, 0.025, 0.05\}.
\]
Each setting is evaluated using 3 random seeds and 40 episodes per seed--$\varepsilon$ pair, resulting in 600 episodes per configuration and 115{,}200 total evaluation episodes.

We report normalized AUC--$\varepsilon$, defined as the area under the return--$\varepsilon$ curve divided by the corresponding clean unattacked baseline. A value of 1.0 indicates clean-baseline performance, values below 1.0 indicate degradation, and values above 1.0 indicate that the evaluated perturbation or defense improves the measured return relative to the clean baseline.

%%%%%%%%%%%%%%%%%%%%%%%%%%%%%%%%%%%%%%%%%%%%%%%%%%%%%%%%%%

\subsection{Benchmark Results}

Table~\ref{tab:auc} reports the normalized AUC--$\varepsilon$ values across all evaluated attack--defense combinations. The results reveal substantial differences in robustness behavior across environments, algorithms, attacks, and defenses.

\begin{table*}[t]
\centering
\caption{Normalised AUC--$\varepsilon$ across all attack\,$\times$\,defense combinations (AUC of attacked run $\div$ AUC of clean unattacked run; 1.0 = clean baseline). Values $>1$ indicate defense raises above-baseline performance; values $<0$ indicate defense alone degrades more than any attack. $\ddag$: Lunar PPO operates near a random baseline; best defense per row in \textbf{bold}.}
\label{tab:auc}
\resizebox{\linewidth}{!}{%
\begin{tabular}{llcccccccc}
\toprule
\textbf{Env} & \textbf{Algo} & \textbf{Atk} & \textbf{None} & \textbf{Smooth} & \textbf{Median} & \textbf{Gauss} & \textbf{FeatSq} & \textbf{OutClip} & \textbf{AdvDet} \\
\midrule
\multirow{9}{*}{Lunar}
 & \multirow{3}{*}{DQN} & FGSM & 0.793 & \textbf{0.805} & 0.794 & 0.728 & 0.547 & $-$0.453 & $-$2.018 \\
 &                      & PGD  & 0.848 & \textbf{0.871} & 0.847 & 0.804 & 0.429 & $-$0.467 & $-$2.019 \\
 &                      & JSMA & \textbf{0.920} & 0.885 & 0.886 & 0.813 & 0.431 & $-$0.499 & $-$2.027 \\
\cmidrule{2-10}
 & \multirow{3}{*}{PPO$^\ddag$} & FGSM & 1.282 & 0.825 & 1.468 & 1.293 & \textbf{1.510} & $-$16.6 & $-$11.7 \\
 &                      & PGD  & 1.428 & 1.316 & 1.426 & \textbf{1.884} & 1.675 & $-$16.4 & $-$11.7 \\
 &                      & JSMA & 1.471 & 1.299 & 1.474 & 1.651 & \textbf{1.762} & $-$16.7 & $-$11.7 \\
\cmidrule{2-10}
 & \multirow{3}{*}{SAC} & FGSM & 0.967 & \textbf{0.970} & 0.968 & 0.827 & 0.387 & $-$0.862 & $-$0.633 \\
 &                      & PGD  & 0.590 & \textbf{0.752} & 0.656 & 0.556 & 0.346 & $-$0.846 & $-$0.634 \\
 &                      & JSMA & 0.802 & \textbf{0.837} & 0.819 & 0.663 & 0.285 & $-$0.851 & $-$0.737 \\
\midrule
\multirow{9}{*}{Hwy}
 & \multirow{3}{*}{DQN} & FGSM & 0.969 & 0.964 & 0.953 & 0.965 & 0.933 & \textbf{0.983} & 0.979 \\
 &                      & PGD  & 0.989 & \textbf{0.993} & 0.979 & 0.987 & 0.911 & 0.988 & 0.979 \\
 &                      & JSMA & \textbf{1.001} & 0.993 & 0.989 & 0.998 & 0.938 & 0.984 & 0.979 \\
\cmidrule{2-10}
 & \multirow{3}{*}{PPO} & FGSM & \textbf{1.000} & \textbf{1.000} & \textbf{1.000} & \textbf{1.000} & \textbf{1.000} & \textbf{1.000} & \textbf{1.000} \\
 &                      & PGD  & \textbf{1.000} & \textbf{1.000} & \textbf{1.000} & \textbf{1.000} & \textbf{1.000} & \textbf{1.000} & \textbf{1.000} \\
 &                      & JSMA & \textbf{1.000} & \textbf{1.000} & \textbf{1.000} & \textbf{1.000} & \textbf{1.000} & \textbf{1.000} & \textbf{1.000} \\
\cmidrule{2-10}
 & \multirow{3}{*}{SAC} & FGSM & \textbf{1.002} & 0.541 & 0.530 & 1.001 & 0.987 & 0.318 & 0.153 \\
 &                      & PGD  & \textbf{1.004} & 0.530 & 0.546 & 1.003 & 0.979 & 0.302 & 0.170 \\
 &                      & JSMA & \textbf{1.003} & 0.547 & 0.549 & 1.002 & 0.982 & 0.314 & 0.166 \\
\bottomrule
\end{tabular}%
}
\end{table*}

%%%%%%%%%%%%%%%%%%%%%%%%%%%%%%%%%%%%%%%%%%%%%%%%%%%%%%%%%%

\subsection{Analysis of Attack Sensitivity}

On LunarLander, DQN exhibits clear degradation under adversarial perturbations. Its normalized AUC decreases to 0.793 under FGSM and 0.848 under PGD. This confirms that single-step and iterative perturbations can meaningfully destabilize a discrete-action value function. FGSM produces stronger degradation than PGD in this setting, suggesting that even a single gradient step is sufficient to move the observation across sensitive decision regions.

SAC is particularly vulnerable to PGD on LunarLander. The undefended PGD AUC drops to 0.590, which is the lowest undefended Lunar result. This behavior is consistent with the continuous and differentiable nature of actor-critic policies, where iterative gradient-based perturbations can exploit smooth policy gradients to construct damaging adversarial directions.

PPO behaves differently from DQN and SAC. Its AUC values exceed 1.0 under several attacks, but this should not be interpreted as genuine robustness. The trained PPO policy operates near a weak baseline on LunarLander, and small perturbations occasionally act as implicit noise that improves the measured return before performance eventually collapses at larger perturbation magnitudes.

JSMA is the weakest attack on LunarLander. Since it perturbs only the top-$k=2$ features in an 8-dimensional observation space, its adversarial coverage is limited compared with FGSM and PGD. For DQN, the undefended JSMA AUC reaches 0.920, which is substantially higher than the corresponding FGSM and PGD results.

%%%%%%%%%%%%%%%%%%%%%%%%%%%%%%%%%%%%%%%%%%%%%%%%%%%%%%%%%%

\subsection{Analysis of Defense Effectiveness}

Temporal smoothing is the most reliable defense on LunarLander. For DQN under FGSM, smoothing improves AUC from 0.793 to 0.805. More importantly, for SAC under PGD, the hardest Lunar setting, smoothing improves AUC from 0.590 to 0.752. This corresponds to a relative recovery of approximately 27\%. The result suggests that temporal averaging can partially suppress correlated adversarial perturbations while preserving task-relevant dynamics.

Median smoothing provides weaker improvements than mean smoothing in most settings. This indicates that the evaluated perturbations are not isolated temporal outliers, but rather structured perturbations that remain correlated across time. In such cases, averaging can be more effective than outlier rejection.

Gaussian noise provides mixed results. It preserves Highway SAC performance and achieves AUC values close to the undefended baseline, but it only partially mitigates Lunar attacks. With $\sigma=0.01$, the injected noise is insufficient to dominate larger adversarial perturbation budgets.

Feature squeezing is generally harmful in LunarLander. For DQN under FGSM, AUC decreases from 0.793 without defense to 0.547 with feature squeezing. This suggests that quantization error can distort low-dimensional control observations more severely than the adversarial perturbation itself. Feature squeezing may therefore be more suitable for high-dimensional image observations than for compact continuous-state control tasks.

%%%%%%%%%%%%%%%%%%%%%%%%%%%%%%%%%%%%%%%%%%%%%%%%%%%%%%%%%%

\subsection{Defense-Induced Failure Modes}

% ^One of the most important findings of the benchmark is that several defenses degrade performance more severely than the attacks they are designed to mitigate. OutlierClip and AdversarialDetector perform poorly across nearly all Lunar settings, producing strongly negative normalized AUC values. For example, Lunar DQN under FGSM drops to $-0.453$ with OutlierClip and $-2.018$ with AdversarialDetector.

% The failure of detection-based methods is especially visible on Highway SAC. Although the undefended SAC policy is essentially attack-robust, with AUC values above 1.0 under all attacks, stateful defenses cause severe performance collapse. Under FGSM, Highway SAC drops from 1.002 without defense to 0.318 with OutlierClip and 0.153 with AdversarialDetector. Similar failures occur under PGD and JSMA.

% These results demonstrate that defenses are not universally beneficial. In particular, stateful defenses that buffer, clip, or replace observations can disrupt the temporal structure required by continuous-action policies. In Highway SAC, smoothing and detection mechanisms likely distort velocity and gap-related features that are essential for stable lane-change and acceleration decisions.

An important finding is that several defenses degrade performance more than the attacks they are intended to mitigate. OutlierClip and AdversarialDetector consistently perform poorly in LunarLander and substantially reduce the performance of Highway SAC, despite the undefended SAC policy being largely robust to adversarial perturbations. These results indicate that stateful defenses can introduce harmful observation distortions that disrupt policy behavior, particularly for continuous-action agents. Consequently, defense effectiveness must be evaluated not only by attack mitigation capability but also by its impact on nominal policy performance.

%%%%%%%%%%%%%%%%%%%%%%%%%%%%%%%%%%%%%%%%%%%%%%%%%%%%%%%%%%

\subsection{Cross-Environment Robustness Patterns}

The benchmark reveals significant environment-dependent robustness behavior. While LunarLander is highly vulnerable to adversarial perturbations and benefits from smoothing-based defenses, Highway-v0 remains largely robust to attacks. However, Highway SAC exhibits substantial performance degradation when defenses introduce observation distortions. These results demonstrate that both attack impact and defense effectiveness depend strongly on the environment and learning algorithm, emphasizing the need for multi-environment robustness evaluation.

\section{Conclusion}
\label{sec:conclusion}

This paper presented \textbf{RoAd-RL}, a unified framework for standardized adversarial robustness evaluation in reinforcement learning. By introducing modular abstractions for policies, attacks, defenses, and metrics, the framework enables reproducible, extensible, and fair robustness benchmarking across diverse environments and learning algorithms.

A comprehensive benchmark involving DQN, PPO, and SAC agents under multiple attack--defense configurations demonstrated that adversarial vulnerability and defense effectiveness are highly dependent on both the environment and the underlying policy. The results further showed that some commonly adopted defenses can significantly degrade nominal performance, highlighting the importance of systematic and reproducible robustness evaluation.

Beyond providing a software framework, RoAd-RL establishes a practical benchmarking infrastructure for analyzing attack--defense interactions, reproducing experimental results, and developing comparable robustness baselines. Future work will extend the framework to support black-box attacks, adversarial training, certified robustness techniques, and multi-agent reinforcement learning environments.

%%%%%%%%%%%%%%%%%%%%%%%%%%%%%%%%%%%%%%%%%%%%%%%%%%%%%%%%%%
\bibliographystyle{IEEEtran}
\bibliography{mybibfile}

% Generated by IEEEtran.bst, version: 1.14 (2015/08/26)
\begin{thebibliography}{10}
\providecommand{\url}[1]{#1}
\csname url@samestyle\endcsname
\providecommand{\newblock}{\relax}
\providecommand{\bibinfo}[2]{#2}
\providecommand{\BIBentrySTDinterwordspacing}{\spaceskip=0pt\relax}
\providecommand{\BIBentryALTinterwordstretchfactor}{4}
\providecommand{\BIBentryALTinterwordspacing}{\spaceskip=\fontdimen2\font plus
\BIBentryALTinterwordstretchfactor\fontdimen3\font minus \fontdimen4\font\relax}
\providecommand{\BIBforeignlanguage}[2]{{%
\expandafter\ifx\csname l@#1\endcsname\relax
\typeout{** WARNING: IEEEtran.bst: No hyphenation pattern has been}%
\typeout{** loaded for the language `#1'. Using the pattern for}%
\typeout{** the default language instead.}%
\else
\language=\csname l@#1\endcsname
\fi
#2}}
\providecommand{\BIBdecl}{\relax}
\BIBdecl

\bibitem{mohan2026toward}
A.~Mohan and T.~Sch{\"o}n, ``Toward robust agents: A survey of adversarial attacks and defenses in deep reinforcement learning,'' \emph{IEEE Access}, 2026.

\bibitem{karpenahalli2025evolution}
C.~Karpenahalli~Ramakrishna, A.~Mohan, Z.~Zeinaly, and L.~Belzner, ``The evolution of criticality in deep reinforcement learning,'' in \emph{Proceedings of the 17th International Conference on Agents and Artificial Intelligence (ICAART 2025)-Volume 3}.\hskip 1em plus 0.5em minus 0.4em\relax SciTePress, 2025, pp. 217--224.

\bibitem{huang2017adversarial}
\BIBentryALTinterwordspacing
S.~Huang, N.~Papernot, I.~Goodfellow, Y.~Duan, and P.~Abbeel, ``Adversarial attacks on neural network policies,'' 2017. [Online]. Available: \url{https://arxiv.org/abs/1702.02284}
\BIBentrySTDinterwordspacing

\bibitem{zhang2020robust}
H.~Zhang, H.~Chen, C.~Xiao, B.~Li, M.~Liu, D.~Boning, and C.-J. Hsieh, ``Robust deep reinforcement learning against adversarial perturbations on state observations,'' in \emph{Advances in Neural Information Processing Systems}, vol.~33, 2020, pp. 21\,024--21\,037.

\bibitem{mohan2025advancing}
A.~Mohan, D.~R{\"o}{\ss}le, D.~Cremers, and T.~Sch{\"o}n, ``Advancing robustness in deep reinforcement learning with an ensemble defense approach,'' \emph{arXiv preprint arXiv:2507.17070}, 2025.

\bibitem{mohan2026real}
A.~Mohan, X.~Xie, V.~T. Sambandham, and T.~Sch{\"o}n, ``Real-time evaluation of autonomous systems under adversarial attacks,'' \emph{arXiv preprint arXiv:2605.03491}, 2026.

\bibitem{goodfellow2014explaining}
I.~J. Goodfellow, J.~Shlens, and C.~Szegedy, ``Explaining and harnessing adversarial examples,'' \emph{arXiv preprint arXiv:1412.6572}, 2014.

\bibitem{madry2017towards}
A.~Madry, A.~Makelov, L.~Schmidt, D.~Tsipras, and A.~Vladu, ``Towards deep learning models resistant to adversarial attacks,'' in \emph{International Conference on Learning Representations}, 2018.

\bibitem{papernot2016crafting}
N.~Papernot, P.~McDaniel, S.~Jha, M.~Fredrikson, Z.~B. Celik, and A.~Swami, ``The limitations of deep learning in adversarial settings,'' in \emph{2016 IEEE European Symposium on Security and Privacy}.\hskip 1em plus 0.5em minus 0.4em\relax IEEE, 2016, pp. 372--387.

\bibitem{sun2020stealthy}
J.~Sun, T.~Zhang, X.~Xie, L.~Ma, Y.~Zheng, K.~Chen, and Y.~Liu, ``Stealthy and efficient adversarial attacks against deep reinforcement learning,'' in \emph{Proceedings of the AAAI Conference on Artificial Intelligence}, vol.~34, no.~04, 2020, pp. 5883--5891.

\bibitem{xu2017feature}
W.~Xu, D.~Evans, and Y.~Qi, ``Feature squeezing: Detecting adversarial examples in deep neural networks,'' in \emph{Network and Distributed System Security Symposium}, 2018.

\bibitem{cohen2019certified}
J.~Cohen, E.~Rosenfeld, and Z.~Kolter, ``Certified adversarial robustness via randomized smoothing,'' in \emph{International Conference on Machine Learning}.\hskip 1em plus 0.5em minus 0.4em\relax PMLR, 2019, pp. 1310--1320.

\bibitem{croce2020robustbench}
\BIBentryALTinterwordspacing
F.~Croce, M.~Andriushchenko, V.~Sehwag, E.~Debenedetti, N.~Flammarion, M.~Chiang, P.~Mittal, and M.~Hein, ``Robustbench: a standardized adversarial robustness benchmark,'' 2021. [Online]. Available: \url{https://arxiv.org/abs/2010.09670}
\BIBentrySTDinterwordspacing

\bibitem{rauber2017foolbox}
J.~Rauber, W.~Brendel, and M.~Bethge, ``Foolbox: A {P}ython toolbox to benchmark the robustness of machine learning models,'' in \emph{ICML 2017 Workshop on Visualization for Deep Learning}, 2017.

\bibitem{papernot2018cleverhans}
\BIBentryALTinterwordspacing
N.~Papernot, F.~Faghri, N.~Carlini, I.~Goodfellow, R.~Feinman, A.~Kurakin, C.~Xie, Y.~Sharma, T.~Brown, A.~Roy, A.~Matyasko, V.~Behzadan, K.~Hambardzumyan, Z.~Zhang, Y.-L. Juang, Z.~Li, R.~Sheatsley, A.~Garg, J.~Uesato, W.~Gierke, Y.~Dong, D.~Berthelot, P.~Hendricks, J.~Rauber, R.~Long, and P.~McDaniel, ``Technical report on the cleverhans v2.1.0 adversarial examples library,'' 2018. [Online]. Available: \url{https://arxiv.org/abs/1610.00768}
\BIBentrySTDinterwordspacing

\bibitem{raffin2021stable}
A.~Raffin, A.~Hill, A.~Gleave, A.~Kanervisto, M.~Ernestus, and N.~Dormann, ``Stable-{B}aselines3: Reliable reinforcement learning implementations,'' \emph{Journal of Machine Learning Research}, vol.~22, no. 268, pp. 1--8, 2021.

\bibitem{towers2024gymnasium}
M.~Towers, A.~Kwiatkowski, J.~Terry, J.~U. Balis, G.~De~Cola, T.~Deleu, M.~Goul{\~a}o, A.~Kallinteris, M.~Krimmel, A.~KG \emph{et~al.}, ``Gymnasium: A standard interface for reinforcement learning environments,'' \emph{arXiv preprint arXiv:2407.17032}, 2024.

\bibitem{pattanaik2018robust}
A.~Pattanaik, Z.~Tang, S.~Liu, G.~Bommannan, and G.~Chowdhary, ``Robust deep reinforcement learning with adversarial attacks,'' in \emph{Proceedings of the 17th International Conference on Autonomous Agents and MultiAgent Systems}, 2018, pp. 2040--2042.

\bibitem{lin2017tactics}
\BIBentryALTinterwordspacing
Y.-C. Lin, Z.-W. Hong, Y.-H. Liao, M.-L. Shih, M.-Y. Liu, and M.~Sun, ``Tactics of adversarial attack on deep reinforcement learning agents,'' 2019. [Online]. Available: \url{https://arxiv.org/abs/1703.06748}
\BIBentrySTDinterwordspacing

\bibitem{lecuyer2019certified}
M.~Lecuyer, V.~Atlidakis, R.~Geambasu, D.~Hsu, and S.~Jana, ``Certified robustness to adversarial examples with differential privacy,'' in \emph{2019 IEEE Symposium on Security and Privacy}.\hskip 1em plus 0.5em minus 0.4em\relax IEEE, 2019, pp. 656--672.

\bibitem{sekaran2025urbaning}
K.~C. Sekaran, M.~Geisler, D.~R{\"o}{\ss}le, A.~Mohan, D.~Cremers, W.~Utschick, M.~Botsch, W.~Huber, and T.~Sch{\"o}n, ``Urbaning-v2x: A large-scale multi-vehicle, multi-infrastructure dataset across multiple intersections for cooperative perception,'' \emph{arXiv preprint arXiv:2510.23478}, 2025.

\bibitem{leurent2018environment}
E.~Leurent, ``An environment for autonomous driving decision-making,'' \url{https://github.com/eleurent/highway-env}, 2018.

\bibitem{rossle2026driving}
D.~R{\"o}{\ss}le, X.~Xie, A.~Mohan, V.~T. Sambandham, D.~Cremers, and T.~Sch{\"o}n, ``Driving: A large-scale multimodal driving dataset with full digital twin integration,'' \emph{arXiv preprint arXiv:2601.15260}, 2026.

\bibitem{mnih2015human}
V.~Mnih, K.~Kavukcuoglu, D.~Silver, A.~A. Rusu, J.~Veness, M.~G. Bellemare, A.~Graves, M.~Riedmiller, A.~K. Fidjeland, G.~Ostrovski \emph{et~al.}, ``Human-level control through deep reinforcement learning,'' \emph{Nature}, vol. 518, no. 7540, pp. 529--533, 2015.

\bibitem{schulman2017proximal}
J.~Schulman, F.~Wolski, P.~Dhariwal, A.~Radford, and O.~Klimov, ``Proximal policy optimization algorithms,'' in \emph{arXiv preprint arXiv:1707.06347}, 2017.

\bibitem{haarnoja2018soft}
T.~Haarnoja, A.~Zhou, P.~Abbeel, and S.~Levine, ``Soft actor-critic: Off-policy maximum entropy deep reinforcement learning with a stochastic actor,'' in \emph{International Conference on Machine Learning}.\hskip 1em plus 0.5em minus 0.4em\relax PMLR, 2018, pp. 1861--1870.

\end{thebibliography}

\end{document}